\pdfoutput=1

\documentclass[11pt]{article}

\usepackage[preprint]{acl}

\usepackage{times}
\usepackage{latexsym}

\usepackage[T1]{fontenc}

\usepackage[utf8]{inputenc}

\usepackage{microtype}

\usepackage{inconsolata}
\usepackage[percent]{overpic}
\usepackage{stmaryrd}
\usepackage{amsfonts}
\usepackage{amssymb}
\usepackage{amsmath}
\usepackage{bbm}
\usepackage{multirow,tabularx}
\usepackage{xspace}
\usepackage{todonotes}
\usepackage{booktabs}
\usepackage{subcaption}
\usepackage{algorithm}
\usepackage{algpseudocode}
\usepackage{marvosym}
\usepackage{hyperref}
\usepackage{authblk}  

\usepackage{amsmath,amsfonts,bm}

\def\eqref#1{equation~\ref{#1}}

\def\1{\bm{1}}

\def\mW{{\bm{W}}}

\DeclareMathAlphabet{\mathsfit}{\encodingdefault}{\sfdefault}{m}{sl}
\SetMathAlphabet{\mathsfit}{bold}{\encodingdefault}{\sfdefault}{bx}{n}

\newcommand{\fedlearn}{\text{FL}\xspace}

\newcommand{\sst}{\text{SST-2}\xspace}
\newcommand{\cola}{\text{CoLA}\xspace}
\newcommand{\rotten}{\text{Rotten Tomatoes}\xspace}

\newcommand{\alGrads}{$\Delta \mW_{all}$\xspace}
\newcommand{\alTF}{$\Delta \mW_{T}$\xspace}

\newcommand{\iT}{$\Delta \mW_{t,i}$\xspace}

\newcommand{\firstTF}{$\Delta \mW_{t,1}$\xspace}

\newcommand{\firstQ}{$\Delta \mW_{q,1}$\xspace}
\newcommand{\firstK}{$\Delta \mW_{k,1}$\xspace}
\newcommand{\firstV}{$\Delta \mW_{v,1}$\xspace}
\newcommand{\firstAout}{$\Delta \mW_{o,1}$\xspace}
\newcommand{\firstFin}{$\Delta \mW_{f,1}$\xspace}
\newcommand{\firstFout}{$\Delta \mW_{p,1}$\xspace}

\usepackage{ntheorem}

\theoremstyle{nonumberplain}

\def\eg{{\em e.g.,}\xspace}
\def\ie{{\em i.e.,}\xspace}

\title{Seeing the Forest through the Trees: Data Leakage \\ from Partial Transformer Gradients}

\author[1]{Weijun Li}
\author[1]{Qiongkai Xu}
\author[1]{Mark Dras}
\affil[1]{School of Computing, FSE, Macquarie University, Sydney, Australia}
\affil[1]{\texttt{weijun.li1@hdr.mq.edu.au, \{qiongkai.xu, mark.dras\}@mq.edu.au}}

\begin{document}
\maketitle
\begin{abstract}
Recent studies have shown that distributed machine learning is vulnerable to gradient inversion attacks, where private training data can be reconstructed by analyzing the gradients of the models shared in training. Previous attacks established that such reconstructions are possible using gradients from all parameters in the entire models. However, we hypothesize that most of the involved modules, or even their sub-modules, are at risk of training data leakage, and we validate such vulnerabilities in various intermediate layers of language models. Our extensive experiments reveal that gradients from a single Transformer layer, or even a single linear component with 0.54\% parameters, are susceptible to training data leakage. Additionally, we show that applying differential privacy on gradients during training offers limited protection against the novel vulnerability of data disclosure.\footnote{Code available at: \url{https://github.com/weijun-l/partial-gradients-leakage}.}
\end{abstract}

\section{Introduction}
As the requirement for training machine learning models on large-scale and diverse datasets intensifies, distributed learning frameworks have risen as an effective solution that balances both the need for intensive computation and critical privacy concerns among edge users. As a prime example, Federated Learning (\fedlearn)~\citep{mcmahan2016federated} preserves the privacy of participants by retaining each client's data on their own devices, while only exchanging essential information, such as model parameters and updated gradients. Nonetheless, recent research~\citep{zhu2019deep, dang2021revealing, balunovic2022lamp} has demonstrated the possibility of reconstructing client data by a curious-but-honest server or clients who have access to the corresponding gradient information.

\begin{figure}[ht]
  \centering
  \begin{overpic}[width=\linewidth]{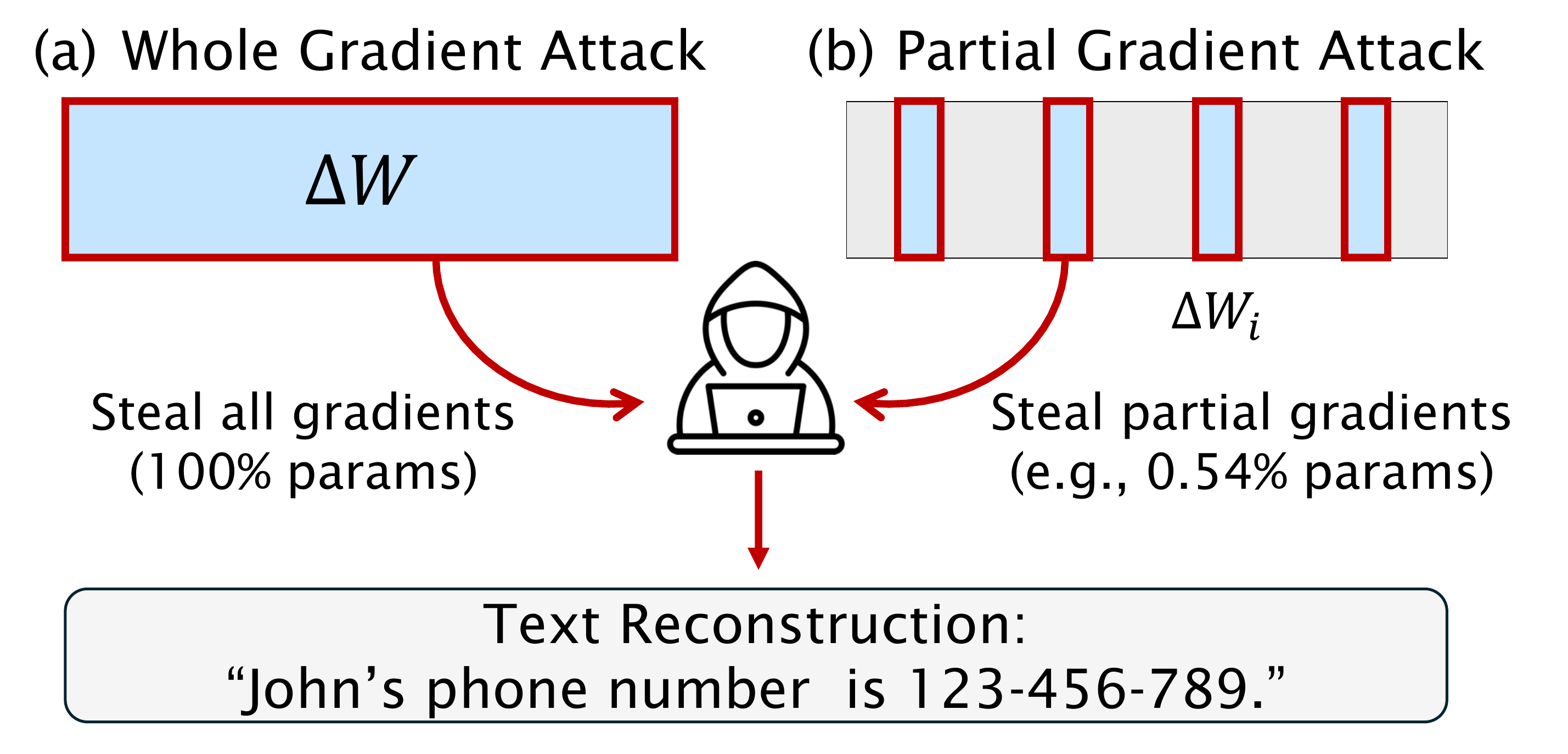}
    \put(2,12){\footnotesize~\cite{balunovic2022lamp}}
    \put(70,12){\footnotesize{(Our work)}}
  \end{overpic}
  \caption{To reconstruct training data, prior attacks (a) typically require access to gradients from the whole model, while our attack (b) uses partial model gradients.}
  \label{fig:teaser}
\end{figure}

Specifically, two types of methods have been proposed to extract private textual training data: (i) gradient matching method~\citep{zhu2019deep} align gradients from the presumed data with the monitored gradients; and (ii) analytical reconstruction techniques~\citep{dang2021revealing,gupta2022recovering} deduce the used tokens by analyzing the gradient patterns, such as the presence of non-zero embedding gradients that are correlated to those used tokens. 

This study examines whether models are more vulnerable than realized by framing a \textbf{research question}: Can private training data be reconstructed using gradients from partial intermediate Transformer modules? This setting is motivated by several realistic scenarios. First, Parameter Freezing~\cite{gupta2022recovering} offers a straightforward defense against reconstruction attacks targeting specific layers, \eg the embedding. Second, layer-wise training strategies are adopted to meet the diverse needs of: (i) transfer learning for domain adaptation~\citep{chen2020fedhealth, saha2021federated}, (ii) personalized federated learning for managing heterogeneous data~\citep{mei2021fedvf} and (iii) enhancing communication efficiency~\citep{lee2023layer}. Generally, existing attacks require access to gradients from either (i) all deep learning modules or (ii) the word embedding/last linear layer; given the above defenses, this is not always practical. 

In this work, we challenge the premise that gradients from all layers are necessary to reconstruct training data. We demonstrate the feasibility of reconstructing text data from gradients of varying granularities, ranging from multiple Transformer layers down to a single layer, or even its single linear component (\eg individual Attention Query, Key modules), as depicted in Figure~\ref{fig:teaser}. Additionally, we investigate the impact of differential privacy on gradients~\citep{abadi2016deep} as a defense and find our attacks remain effective, without significant degradation of model performance. Our study motivates further research into more effective defense mechanisms for distributed learning applications.

\section{Related Work}

\paragraph{Distributed Learning.} 
Distributed learning, such as Federated Learning (\fedlearn), is a growing field aimed at parallelizing model training for better efficiency and privacy~\citep{gade2018privacy,verbraeken2020survey,froelicher2020scalable}. \fedlearn is highly valuable for privacy preservation by remaining sensitive data local as participants compute gradients on their devices and sharing updates via a central server~\citep{mcmahan2017communication}.

However, the shared gradients introduce an attack surface that allows malicious participants, or the curious-but-honest server, to reconstruct the training data. The gradient assets available to attackers may vary depending on the use of parameter freezing defense~\citep{gupta2022recovering} or layer-wise training strategies~\citep{lee2023layer}. 

\paragraph{Gradient Inversion Attack (GIA).} Recent studies~\citep{zhu2019deep,zhao2020idlg,yang2023gradient,zhang2024revisiting} have investigated data leakage in distributed learning, known as Gradient Inversion Attack~\cite{zhang2022survey,du2024sok}. One strategy is the \textbf{analytical-based} approach, which identifies correlations between the gradients and model parameters to retrieve used training tokens. RLG~\citep{dang2021revealing}, FILM~\citep{gupta2022recovering}, DECEPTICONS~\citep{fowl2022decepticons} and DAGER~\citep{petrov2024dager} have demonstrated the use of gradients from specific layers, such as last linear and embedding layers, for reconstruction. While these works present effective attacks, parameter freezing~\citep{gupta2022recovering} can address threats from specific layers, and the method of DECEPTICONS assumes malicious parameter modification. 

Alternatively, \textbf{optimization-based} method iteratively refines randomly generated data to match real data by minimizing the distance between their gradients across layers. Pioneering works like DLG~\citep{zhu2019deep}, InvertGrad~\citep{geiping2020inverting}, and TAG~\citep{deng2021tag} employed Euclidean (L2), Cosine, and combined Euclidean and Manhattan (L1) distances for data reconstruction. LAMP~\citep{balunovic2022lamp} advances these attacks with selective initialization, embedding regularization loss, and word reordering. \citet{wu2023learning} and \citet{li2023beyond} build on the frameworks of DLG or LAMP while incorporating additional information, such as auxiliary datasets. Our study also adheres to optimization-based strategy but shows that partial gradients alone can reveal private training data. 

Recent work has explored reconstructing private training data from single gradient modules, but with different focuses from ours. \citet{wang2023reconstructing} employs gradient decomposition in a synthetic two-layer network to recover training examples, though applying this method to practical deep networks requires idealized conditions such as transparent parameter manipulation and controlled activations (\eg tanh, sigmoid). \citet{li2023beyond} extends this approach to recover hidden representations from BERT's pooler layer, guiding optimization via the LAMP method that leverages all gradients. DAGER~\citep{petrov2024dager} uses the first two self-attention gradients of transformer-based language models for reconstruction, locating valid tokens by determining whether their embedding vectors lie within the subspace of the first layer's gradient, then combining their embeddings to verify their occurrence with the second layer’s gradient to establish exact sequences. Unlike these methods, our research focuses on identifying vulnerabilities across all intermediate modules to gradient inversion attacks.

\paragraph{Defense against GIA.} To mitigate the risk of GIA, two defense strategies have been explored: i) \textbf{encryption-based} methods, which disguise the real gradients using techniques such as Homomorphic Encryption~\citep{254465} and Multi-Party Computation~\citep{mugunthan2019smpai}, and ii) \textbf{perturbation-based} methods, including gradient pruning~\citep{zhu2019deep}, adding noise through differential privacy~\citep{balunovic2021bayesian}, or through learned perturbations~\citep{sun2021soteria, fan2024guardian}. The former approach incurs additional computational costs~\citep{fan2024guardian}, while the latter may face challenges in achieving a balance in the privacy-utility trade-off. In this work, we compare the vulnerabilities associated with attacking partial gradients versus the previously fully exposed setting under the defense of differential privacy on gradients~\citep{abadi2016deep}.

\section{Data Leakage from Partial Gradients}
In this section, we first introduce the threat model and then present the attack methodology, which enables the use of intermediate gradients in Transformer layers to reconstruct training data.

\paragraph{Threat Model.} The attack operates in distributed training environments where the server distributes initial model weights and clients submit gradients derived from their local data. \textit{Potential attackers}, either participating as clients or as curious-but-honest servers, can access shared parameters and intercept gradient communications. Unlike the \textit{assets} in prior studies that permitted access to full gradients, our research focuses on scenarios where attackers only observe partial gradients. The goal of these attackers is to reconstruct the private text data received by other clients or servers in training.

\paragraph{Attack Strategy.} Inspired by the gradient matching strategy, which involves minimizing the distance between gradients generated from randomly sampled data and the true gradients to iteratively align the dummy input with the real one, we formulate our optimization objective as,
\begin{equation}
\mathcal{L}= \sum_{i \in \mathcal{N}} \sum_{m \in \mathcal{M}} \mathcal{D} (\Delta \mW'_{m, i}, \Delta \mW_{m, i}),
\end{equation}
where:
\begin{itemize}
    \item $\mathcal{N}$ is a non-empty subset of Transformer layers $\{1, 2, 3, \cdots, l\}$ in an $l$-layer network.
    \item $\mathcal{M}$ is a non-empty subset of the available modules $\{q, k, v, o, f, p\}$ within each layer.
\end{itemize}
\begin{table}[ht]
\centering
\resizebox{\columnwidth}{!}{%
\begin{tabular}{llcc}
\toprule
\textbf{Employed Gradients} & \textbf{Notation} & \textbf{\#. Parameters} & \textbf{Used Ratio \%} \\
\midrule
All Layers (Baseline) & $\Delta \mW_{all}$ & 109,483,778 & 100 \\
\midrule
All \textbf{T}ransformer Layers & $\Delta \mW_{T}$ & 85,054,464 & 77.69 \\
\midrule
$i$-th \textbf{T}ransformer Layer & $\Delta \mW_{t,i}$ & 7,087,872 & 6.47 \\
\midrule
$i$-th FFN Out\textbf{p}ut & $\Delta \mW_{p,i}$ & 2,359,296 & 2.15 \\
$i$-th FFN \textbf{F}ully Connected & $\Delta \mW_{f,i}$ & 2,359,296 & 2.15 \\
$i$-th Attention \textbf{O}utput  & $\Delta \mW_{o,i}$ & 589,824 & 0.54 \\
$i$-th Attention \textbf{Q}uery & $\Delta \mW_{q,i}$ & 589,824 & 0.54 \\
$i$-th Attention \textbf{K}ey & $\Delta \mW_{k,i}$ & 589,824 & 0.54 \\
$i$-th Attention \textbf{V}alue & $\Delta \mW_{v,i}$ & 589,824 & 0.54 \\
\bottomrule
\end{tabular}%
}
\caption{Notations of varying gradient modules, and their parameter numbers for a BERT$_{\text{BASE}}$ model.} 
\label{table:param_notation}
\end{table}
We detail these notations and their involved number of parameters in Table~\ref{table:param_notation}. The lowest ratio is 0.54\% for attacking a single linear module. $\Delta \mW$ and $\Delta \mW'$ represent the target gradient and the derived dummy gradient, respectively. $\mathcal{D}$ serves as a distance measurement in optimization. We extend the use of cosine distance, following prior studies~\citep{balunovic2022lamp, geiping2020inverting},
\begin{align}
\mathcal{D}_{\text{cos}} (& \Delta \mW'_{m,i}, \Delta \mW_{m,i}) \nonumber\\
&= 1 - \frac{\Delta \mW'_{m,i} \cdot \Delta \mW_{m,i}}{\|\Delta \mW'_{m,i}\| \cdot \|\Delta \mW_{m,i}\|},
\end{align}
which was reported to be stable and outperform other metrics (\ie L2 and L1).

Targeting different gradient units, our method can construct various specific gradient matching settings, \eg merely involving one Attention Query component~($q$) in the $i$-th Transformer layer, leading to the overall loss objective,
\begin{equation}
\mathcal{L}= \mathcal{D} (\Delta \mW'_{q, i}, \Delta \mW_{q, i}).
\end{equation}

By involving all gradients and assigning normalized equal weights $1/l$, the objective can be specialized to the prior work~\cite{balunovic2022lamp}, \ie
\begin{equation}
\mathcal{L}= \frac{1}{l} \sum_{i=1}^{l} \mathcal{D} (\Delta \mW'_{i}, \Delta \mW_{i}).
\end{equation}

We explore varying degrees of gradient involvement for optimization alignment, from all Transformer layers to a single layer or even an individual linear component, revealing that every single module within the Transformer is vulnerable.

\section{Experiments}
\subsection{Experimental Settings}

\paragraph{Datasets.} We examine reconstruction attacks on three classification datasets: \textbf{\cola}~\citep{warstadt2019neural}, \textbf{\sst}~\citep{socher2013recursive}, and \textbf{\rotten}~\citep{pang2005seeing}, following previous studies~\citep{balunovic2022lamp, li2023beyond} to evaluate reconstruction performance. We conduct experiments on 10 batches randomly sampled from each dataset separately, reporting average results across different test scenarios.

\paragraph{Models.} We conduct experiments on BERT$_{\text{BASE}}$, BERT$_{\text{LARGE}}$~\citep{devlin-etal-2019-bert}, and TinyBERT \citep{jiao2019tinybert}, following previous studies~\citep{deng2021tag, balunovic2022lamp, li2023beyond}. We also adopted a BERT$_{\text{FT}}$ model, which involves fine-tuning BERT$_{\text{BASE}}$ for two epochs before attacks. For the word reordering step after reconstruction, we utilized a customized GPT-2 language model trained by~\citet{guo2021gradient} as an auxiliary tool, same as the setting used in LAMP.

\paragraph{Evaluation Metrics.} We evaluate the efficacy of our attack using \textbf{ROUGE-1}, \textbf{ROUGE-2}, and \textbf{ROUGE-L}~\citep{rouge2004package, lhoest-etal-2021-datasets}, corresponding to unigram, bigram, and longest common sub-sequence, respectively. The F-scores of these metrics are reported.

\paragraph{Attack Setup.} We adapt the open-source implementation of LAMP~\citep{balunovic2022lamp} to serve as both the basis framework and the baseline, as LAMP remains the state-of-the-art method for text reconstruction. We employ identical hyper-parameters as LAMP for fair comparisons. The engineering contribution of our work is that we implemented a gradient extraction process to obtain partial gradients from modules at varying desired granularity.

\paragraph{Defense Setup.} We employ DP-SGD~\citep{abadi2016deep, yousefpour2021opacus}, adjusting the noise multiplier $\sigma$ while maintaining a clipping bound $C$ as $1.0$. To assess noise effects, we train a BERT$_{\text{BASE}}$ model on the \textbf{SST-2} dataset for 2 epochs, evaluating utility changes with the F1-score and MCC~\citep{matthews1975comparison,chicco2020advantages}. We set the delta $\delta$ to $2 \times 10^{-5}$, and explore noise multipliers from 0.01 to 0.5.\footnote{$\delta$ is recommended to be smaller than $1/|D|$~\citep{abadi2016deep}, where $|D|$ is the dataset size.}

\subsection{Results and Analysis}
\begin{figure}[!ht]
  \centering
  \includegraphics[width=\linewidth]{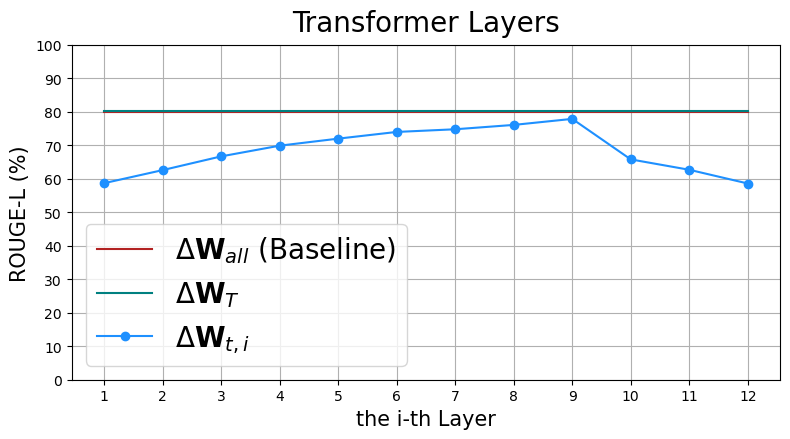}
  \caption{Results across varying \textbf{Transformer} layers.}
  \label{fig:tf_wise}
\end{figure}

\begin{figure}[!ht]
  \centering
  \includegraphics[width=\linewidth]{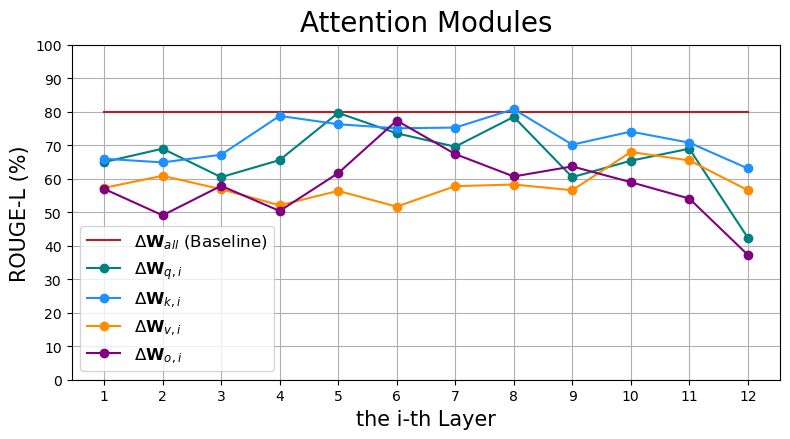}
  \caption{Results across varying \textbf{Attention} Modules.}
  \label{fig:attn_wise}
\end{figure}

\begin{figure}[!ht]
  \centering
  \includegraphics[width=\linewidth]{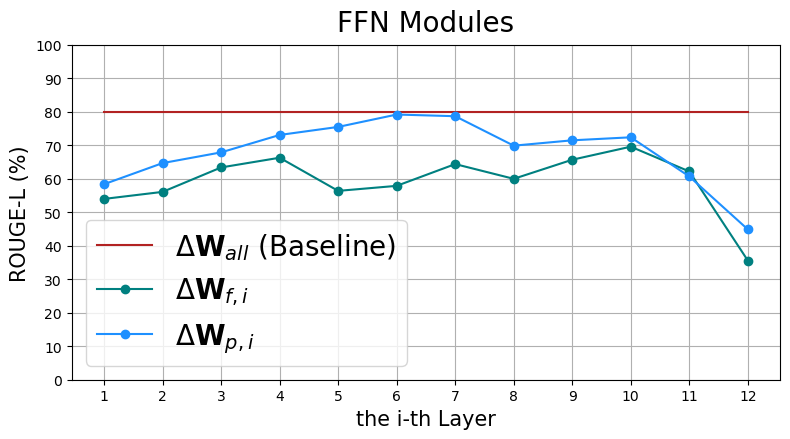}
  \caption{Results across varying \textbf{FFN} Modules.}
  \label{fig:ffn_wise}
\end{figure}

\paragraph{Attack Results.} We present the ROUGE-L scores for experiments on various gradient granularities, including results for Transformer layers in Figure~\ref{fig:tf_wise}, Attention modules across all layers in Figure~\ref{fig:attn_wise}, and FFN modules in Figure~\ref{fig:ffn_wise}. These tests used the BERT$_{\text{BASE}}$ model on the \textbf{\cola} dataset with a batch size of 1. Further results on ROUGE-1 and ROUGE-2 scores for \textbf{\cola} are shown in Figure~\ref{fig:cola_all}, \textbf{\sst} in Figure~\ref{fig:sst_all}, and \textbf{\rotten} in Figure~\ref{fig:rotten_all}. More results on different models and larger batch sizes (B = 2, 4) are provided in Table~\ref{table:more_models} and Table~\ref{table:more_batch_size} in Appendix~\ref{sec:appendix_a}, along with several reconstruction examples presented in Table~\ref{table:example} in Appendix~\ref{sec:appendix_b}.

By inspecting Figure~\ref{fig:tf_wise}, we observe that using gradients from Transformer layers \alTF achieves performance comparable to the baseline, which uses all gradients. Additionally, merely using gradients from a single layer \iT still yields decent attack scores, with layers 6 to 9 achieving results comparable to the baseline while using only 6.47\% of model parameters. These results demonstrate that each layer is vulnerable to the reconstruction attack, with the middle layers hold highest risk.

\begin{table*}[!htbp]
\centering
\resizebox{0.99\linewidth}{!}{
\begin{tabular}{cc|ccc|ccc|ccc|ccc}
\toprule
\multirow{3}{*}{\textbf{Method}} & \multirow{3}{*}{\textbf{Gradient}}  & \multicolumn{3}{c|}{$\sigma = 0.01$} & \multicolumn{3}{c|}{$\sigma = 0.1$} & \multicolumn{3}{c|}{$\sigma = 0.3$} & \multicolumn{3}{c}{$\sigma = 0.5$} \\ \cline{3-14}
 &  &   \multicolumn{3}{c|}{$\varepsilon = 2*10^8$} & \multicolumn{3}{c|}{$\varepsilon = 574.55$} & \multicolumn{3}{c|}{$\varepsilon = 9.38$} & \multicolumn{3}{c}{$\varepsilon = 1.87$} \\ \cline{3-14}
 &  &   R-1 & R-2 & R-L & R-1 & R-2 & R-L & R-1 & R-2 & R-L & R-1 & R-2 & R-L   \\ \bottomrule
LAMP & \alGrads & \textbf{80.6} & 39.0 & \textbf{68.3} & \textbf{79.5} & \textbf{46.3} & \textbf{68.3} & 76.8 & \textbf{38.4} & \textbf{66.8} & 75.0 & 34.6 & 64.9 \\ \bottomrule
\multirow{6}{*}{Ours} & \alTF & 78.3 & \textbf{40.3} & 65.9 & 75.7 & 38.7 & 64.4 & \textbf{77.6} & 36.6 & 66.5 & \textbf{75.9} & \textbf{50.2} & \textbf{70.5} \\ \cline{2-14}
  & \firstTF   & 64.2 & 31.3 & 56.4 & 69.7 & 31.4 & 62.3 & 64.9 & 30.0 & 62.6 & 67.7 & 32.7 & 60.1 \\ \cline{2-14}
 & \firstQ   & 69.8 & 35.8 & 64.5 & 69.3 & 17.6 & 56.0 & 68.0 & 16.0 & 58.4 & 71.1 & 19.5 & 59.4 \\ \cline{2-14}
  & \firstAout   & 52.0 & 10.5 & 48.1 & 52.6 & 12.7 & 47.0 & 49.0 & 1.8 & 43.1 & 46.1 & 12.8 & 43.9 \\ \cline{2-14}
  & \firstFin   & 54.2 & 12.4 & 49.8 & 57.1 & 9.0 & 49.8 & 57.4 & 9.5 & 49.9 & 57.9 & 8.2 & 48.8 \\ \cline{2-14}
  & \firstFout   & 64.3 & 24.8 & 54.2 & 58.3 & 19.9 & 50.6 & 54.7 & 23.2 & 49.3 & 58.7 & 23.3 & 52.9 \\ \bottomrule
   \multicolumn{14}{c}{\textbf{(a) Attack Performance}}  \\ 
   \\[-1.0em]
   \toprule
   \multicolumn{2}{c|}{F1-Score} &   \multicolumn{3}{c|}{$ 0.891$} & \multicolumn{3}{c|}{$0.855$} & \multicolumn{3}{c|}{$0.71$} & \multicolumn{3}{c}{$0.675$} \\ \midrule
   \multicolumn{2}{c|}{MCC} &   \multicolumn{3}{c|}{0.773} & \multicolumn{3}{c|}{0.709} & \multicolumn{3}{c|}{0.285} & \multicolumn{3}{c}{0} \\ \bottomrule
   \multicolumn{14}{c}{\textbf{(b) Model Utility}}
\end{tabular}
}
\caption{Evaluation of differential privacy defense on \sst dataset for varying gradient settings with batch size = 1.}
\label{table:defense}
\end{table*}

Figure~\ref{fig:attn_wise} presents the results of the attack from individual modules in the Attention Blocks across all layers. Most modules facilitate attack performance above 50\%, while the Query and Key modules achieve relatively higher attack performance. Similarly, middle layers achieve the best performance; surprisingly, $\Delta \mW_{k,4}$, $\Delta \mW_{k,8}$, $\Delta \mW_{q,5}$ and $\Delta \mW_{o,6}$ achieve performance almost equivalent to the baseline, while using only 0.54\% of its parameters. Similar results can be observed from FFN modules, as demonstrated in Figure~\ref{fig:ffn_wise}.

\paragraph{Defense Results.} We deploy differential privacy via DP-SGD~\citep{abadi2016deep} to counter data reconstruction attacks. In our \sst dataset experiments, we apply varying noise levels ($\sigma$) from $0.01$ to $0.5$ under privacy budgets ($\epsilon$) of $2 \times 10^8$ to $1.87$. Noise levels above $0.5$ were not explored due to a significant drop in the MCC metric from 0.773 to 0, compromising model utility.  

The results are detailed in Table~\ref{table:defense}. With the increase in noise, we observe a considerable decline in model utility. This observation supports our conclusion that while DP-SGD offers limited protection against the attack, it significantly impacts model utility, as the MCC and F1-Score metrics decreased substantially with a privacy budget $\varepsilon$ of 1.87.

\section{Conclusion}
In this study, we investigate the feasibility of reconstructing training data using partial gradients in a Transformer model. Our extensive experiments demonstrate that all modules within a Transformer are vulnerable to such attacks, leading to a much higher degree of privacy risk than has previously been shown. Our examination of differential privacy as a defense also indicates that it is not sufficient to safeguard private text data from exposure, inspiring further efforts to mitigate these risks.

\section*{Limitations}
We identify several opportunities for further improvement.

We conduct our experiments in the context of classification tasks; however, we propose that the task of language modeling could serve as an additional viable application scenario. Our preliminary validation revealed promising outcomes. However, we did not scale the test volume or involve them due to limitations in computational resources.

We evaluate the defense effectiveness of DP-SGD and find that it is not adequate to mitigate the risk without significant model utility degradation. However, we believe that other techniques, such as Homomorphic Encryption and privacy-preserved multi-party communication, could potentially reduce such a risk of privacy leakage. Nonetheless, these techniques often impose a significant overhead on system communication and substantial computational resources. Therefore, we advocate for further research into the defense strategies that can enhance the system's resilience more efficiently.

In our experiments, we set the batch sizes to $1, 2, 4$ and the average sentence length is less than 25 words, resulting in a total tokens involved being smaller than in generic industrial settings. We consider exploring larger batch sizes and longer sequences as future research directions, which were discussed in the related work, DAGER~\citep{petrov2024dager}.

While our experiments adhered to controlled settings similar to previous work such as LAMP, TAG, DLG—aimed at identifying foundational vulnerabilities in Transformer models, widely used in realistic applications—we also recognize the value of exploring more real-world settings in future work. These could include network delays, asynchronous updates, and heterogeneous data, among others.

\section*{Ethics Statement}
In this study, we delve into the vulnerabilities of every module in Transformer-based models against data reconstruction attacks. Our investigation seeks to evaluate the resilience of cutting-edge Transformer models when they are trained in a distributed learning setting and face malicious reconstruction attacks. Our findings reveal that each component is susceptible to these attacks.

Our research suggests that the risk of data breaches could be more significant than initially estimated, emphasizing the vulnerability of the entire Transformer architecture. We believe it is crucial to disclose such risks to the public, encouraging the research community to take these factors into account when developing secure systems and applications, and to promote further research into effective defense strategies.

\section*{Acknowledgement}
We would like to express our appreciation to the anonymous reviewers for their valuable feedback. This research was undertaken with the assistance of resources from the National Computational Infrastructure (NCI Australia), an NCRIS enabled capability supported by the Australian Government. We also express our gratitude to SoC Incentive Fund, FSE strategic startup grant and HDR Research Project Funding for supporting both travel and research.

\bibliography{custom}

\appendix

\newpage
\onecolumn
\section{Reconstruction Attack Results on Various Configurations}
\label{sec:appendix_a}

We conduct experiments utilizing varying settings on \textit{three datasets}, \ie \cola, \sst and \rotten, under \textit{four variants} of Transformer-based models, \ie BERT$_{\text{BASE}}$, BERT$_{\text{FT}}$, TinyBERT,  BERT$_{\text{LARGE}}$, combined with \textit{different batch sizes} ($B = 1, 2, \text{ and } 4$). The results are presented in the following sections.

\vspace{11pt}

\subsection{Attack Performance on Different Datasets}
\label{sec:sub_appendix_1}
\begin{figure}[!htbp]
  \centering
  \includegraphics[width=\linewidth]{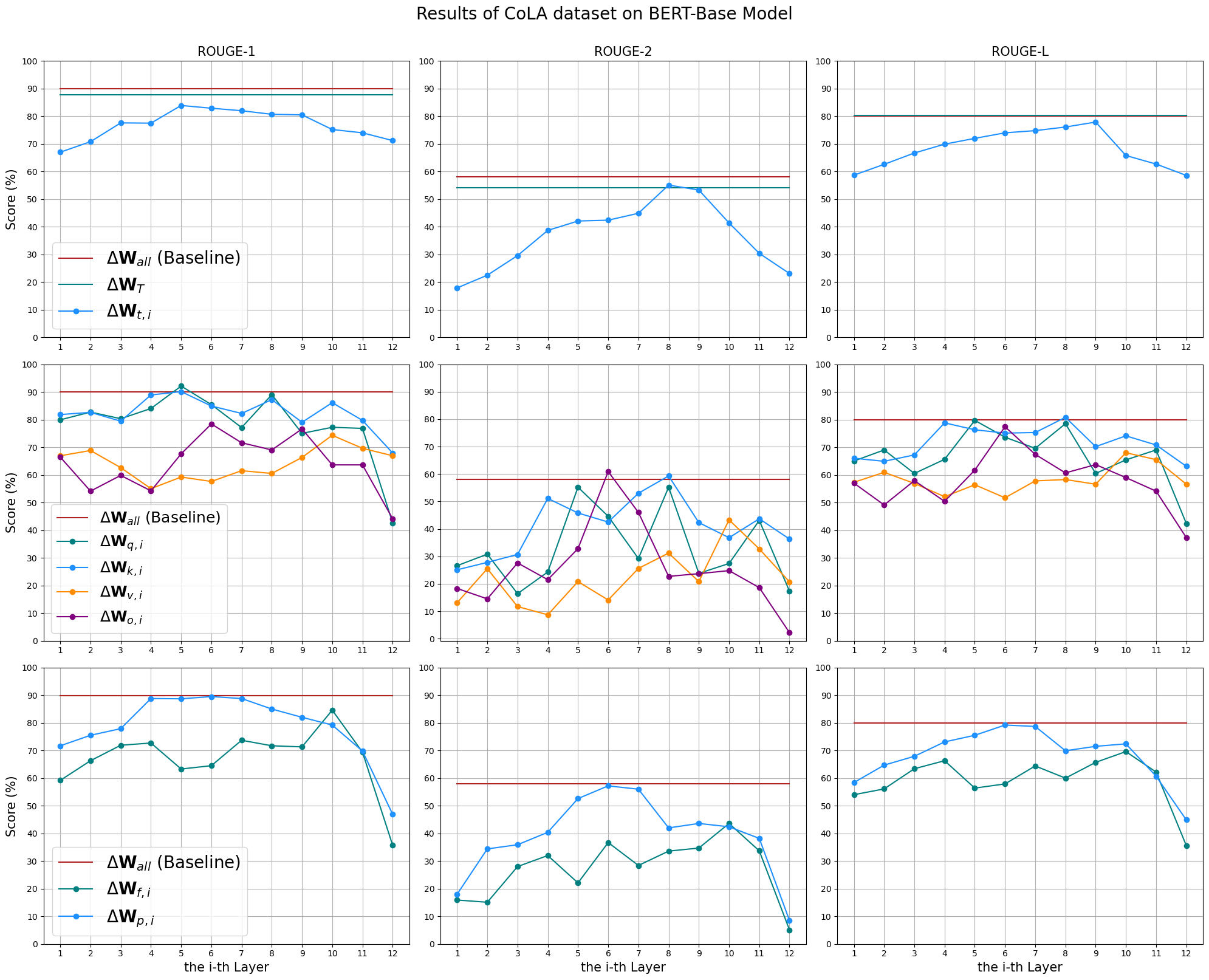}
  \caption{The comparison of reconstruction attacks using different gradient modules on \textbf{\cola} dataset and BERT$_{\text{BASE}}$ model ($B = 1$).}
  \label{fig:cola_all}
\end{figure}

\begin{figure}[!htbp]
  \centering
  \includegraphics[width=\linewidth]{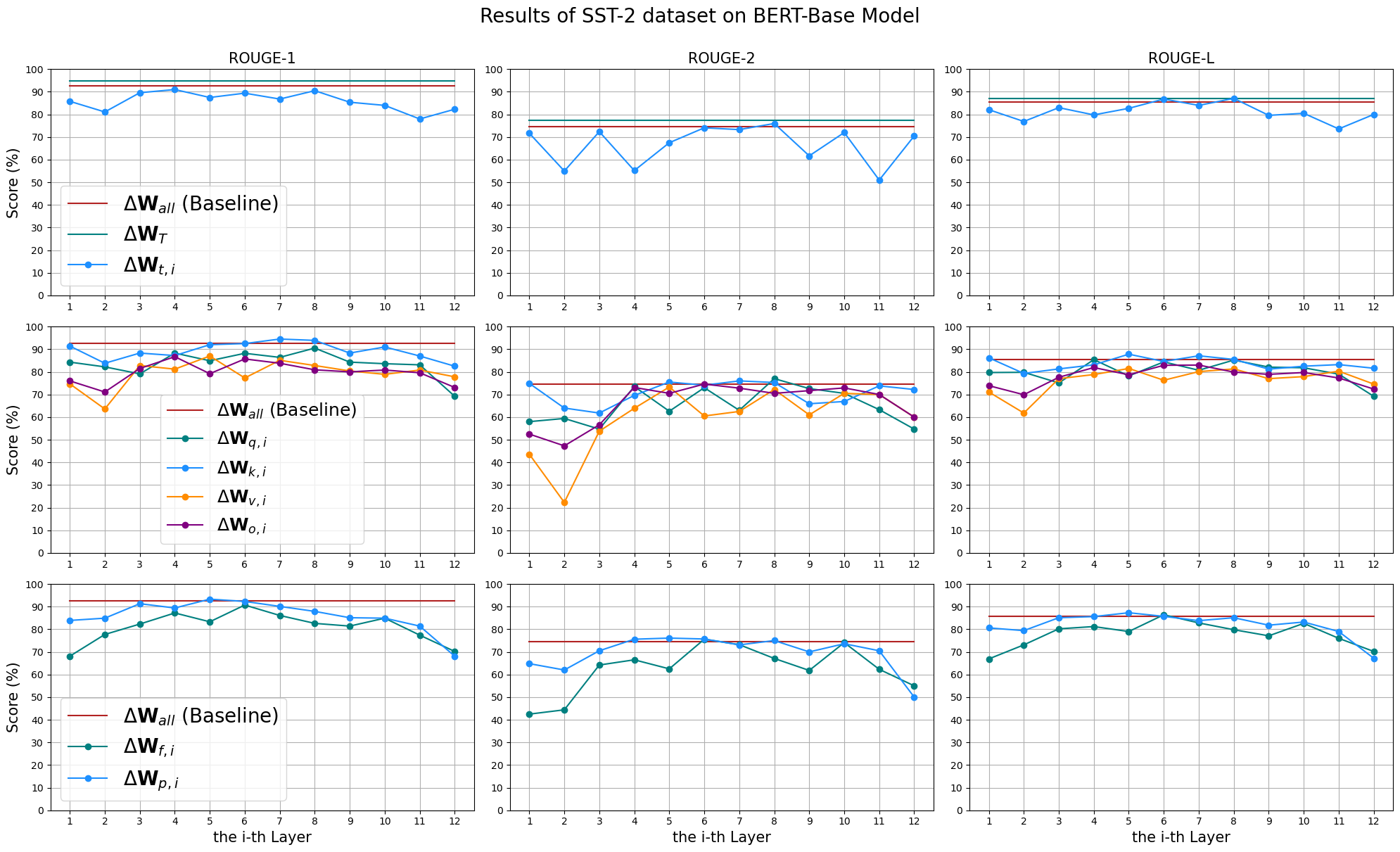}
  \caption{The comparison of reconstruction attacks using different gradient modules on \textbf{\sst} dataset and BERT$_{\text{BASE}}$ model ($B = 1$).}
  \label{fig:sst_all}
\end{figure}

\begin{figure}[!htbp]
  \centering
  \includegraphics[width=\linewidth]{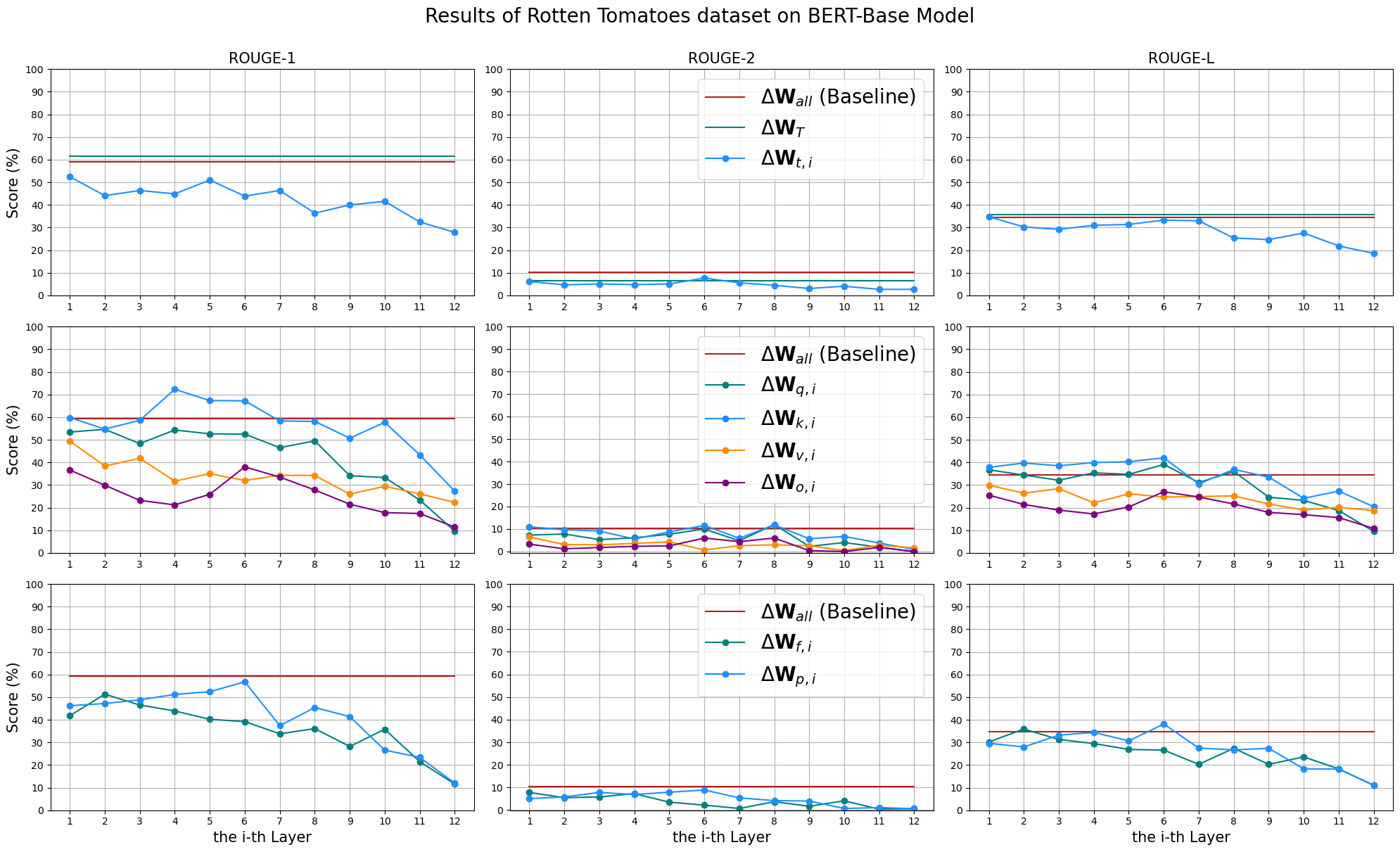}
  \caption{The comparison of reconstruction attacks using different gradient modules on \textbf{\rotten} dataset and BERT$_{\text{BASE}}$ model ($B = 1$).}
  \label{fig:rotten_all}
\end{figure}

We conduct attacks using different gradient modules of BERT$_{\text{BASE}}$ across the three datasets, \textbf{\cola}, \textbf{\sst}, and \textbf{\rotten}. The results are depicted in Figure~\ref{fig:cola_all}, Figure~\ref{fig:sst_all}, and Figure~\ref{fig:rotten_all}. 

We observe three key points: (i) gradients from all Transformer layers perform comparably to the baseline, which uses gradients from all deep learning modules, (ii) gradients from specific modules, such as Attention Query, Attention Key, and FFN Output, can achieve equivalent or superior performance to the baseline, and (iii) gradients from the middle layers outperform those from the shallow or final layers.

\newpage
\subsection{Attack Performance on Different Models}
\label{sec:sub_appendix_2}
We compare reconstruction attacks using four models, and the results are presented in Table~\ref{table:more_models}.
\begin{table*}[!htbp]
\centering
\resizebox{\linewidth}{!}{
\begin{tabular}{cc|ccc|ccc|ccc|ccc}
\toprule
\multirow{2}{*}{\textbf{Method}} & \multirow{2}{*}{\textbf{Gradient}} & \multicolumn{3}{c|}{\textbf{BERT$_\text{BASE}$}} & \multicolumn{3}{c|}{\textbf{BERT$_\text{FT}$}} & \multicolumn{3}{c|}{\textbf{TinyBERT}} & \multicolumn{3}{c}{\textbf{BERT$_\text{LARGE}$}} \\ \cline{3-14}

 &  &  R-1 & R-2 & R-L & R-1 & R-2 & R-L & R-1 & R-2 & R-L & R-1 & R-2 & R-L   \\ \bottomrule
\multicolumn{14}{c}{\textbf{Part A : \cola}}  \\ \toprule
LAMP & \alGrads & \textbf{89.9} & \textbf{58.0} & 80.0 & \textbf{87.8} & \textbf{53.9} & \textbf{78.5} & 95.2 & 67.5 & 86.3 & 90.0 & \textbf{51.6} & 77.3 \\ \midrule
\multirow{8}{*}{Ours} & \alTF & 87.8 & 54.2 & \textbf{80.4} & 82.8 & 38.5 & 71.2 & 97.6 & 59.4 & 85.2 & \textbf{93.6} & 48.9 & \textbf{77.7} \\ \cline{2-14}
  & \firstTF & 67.0 & 17.9 & 58.7 & 73.4 & 28.8 & 61.6 & 93.4 & 47.5 & 79.5 & 84.3 & 34.1 & 70.2 \\ \cline{2-14}
 & \firstQ & 79.9 & 26.6 & 65.0 & 72.1 & 20.8 & 60.1 & 86.0 & 36.2 & 72.1 & 78.7 & 24.9 & 62.2 \\ \cline{2-14}
  & \firstK & 81.8 & 25.1 & 66.0 & 75.9 & 31.1 & 66.4 & 79.5 & 35.7 & 68.7 & 66.9 & 21.7 & 57.0 \\ \cline{2-14}
   & \firstV & 66.9 & 13.1 & 57.3 & 66.5 & 21.0 & 58.2 & 71.7 & 25.6 & 65.8 & 64.4 & 9.9 & 54.3 \\ \cline{2-14}
  & \firstAout & 66.4 & 18.3 & 57.0 & 63.6 & 12.8 & 55.9 & 85.8 & 47.6 & 75.7 & 72.3 & 26.6 & 61.0 \\ \cline{2-14}
  & \firstFin & 59.2 & 15.9 & 54.0 & 61.9 & 12.9 & 53.1 & \textbf{97.8} & 51.0 & 77.0 & 64.2 & 12.0 & 56.4 \\ \cline{2-14}
  & \firstFout & 71.7 & 18.0 & 58.4 & 65.9 & 16.1 & 52.8 & 95.6 & \textbf{68.5} & \textbf{88.1} & 75.6 & 23.2 & 67.7 \\ \bottomrule
 \multicolumn{14}{c}{\textbf{Part B : \sst}}  \\ \toprule
LAMP & \alGrads & 92.6 & 74.5 & 85.6 & \textbf{92.5} & \textbf{70.5} & \textbf{85.7} & 91.5 & 66.0 & 81.3 & \textbf{95.4} & 70.5 & \textbf{85.3} \\ \midrule
\multirow{8}{*}{Ours} & \alTF & \textbf{94.9} & \textbf{77.4} & \textbf{87.1} & 89.0 & 54.3 & 80.1 & \textbf{97.0} & 67.7 & 86.7 & 92.0 & \textbf{73.0} & 83.8 \\ \cline{2-14}
  & \firstTF & 85.8 & 71.7 & 82.1 & 81.4 & 61.6 & 77.6 & 92.2 & 67.4 & 85.6 & 86.8 & 51.8 & 76.6 \\ \cline{2-14}
 & \firstQ & 84.3 & 58.0 & 79.7 & 77.7 & 45.5 & 71.5 & 93.2 & 68.0 & 85.9 & 88.7 & 54.3 & 79.6 \\\cline{2-14}
   & \firstK & 91.4 & 74.9 & 86.2 & 82.9 & 55.9 & 80.5 & 81.4 & 42.8 & 74.2 & 71.4 & 40.0 & 67.8 \\ \cline{2-14}
   & \firstV & 74.7 & 43.6 & 71.1 & 74.9 & 39.7 & 71.2 & 85.5 & 55.7 & 78.0 & 67.9 & 39.4 & 66.9 \\ \cline{2-14}
  & \firstAout & 76.0 & 52.5 & 73.9 & 73.4 & 36.5 & 68.9 & 86.9 & 55.5 & 79.8 & 81.5 & 52.9 & 77.0 \\ \cline{2-14}
  & \firstFin & 68.1 & 42.5 & 66.9 & 72.7 & 29.4 & 67.7 & 96.8 & 68.0 & \textbf{87.2} & 74.1 & 26.1 & 66.1 \\\cline{2-14}
  & \firstFout & 83.9 & 64.8 & 80.6 & 79.9 & 43.0 & 72.1 & 96.3 & \textbf{68.2} & 86.0 & 79.9 & 34.4 & 70.1 \\\bottomrule
 \multicolumn{14}{c}{\textbf{Part C : \rotten}}  \\ \toprule
LAMP & \alGrads & 59.2 & \textbf{10.3} & 34.6 & 63.0 & 6.6 & 37.5 & 74.1 & 26.3 & 52.0 & \textbf{70.9} & 9.8 & 40.8 \\ \midrule
\multirow{8}{*}{Ours} & \alTF & \textbf{61.6} & 6.4 & 35.9 & \textbf{66.8} & \textbf{15.8} & \textbf{43.5} & 74.5 & 17.5 & 48.0 & 68.5 & 7.5 & 34.8 \\ \cline{2-14}
  & \firstTF & 52.5 & 6.2 & 34.8 & 44.6 & 6.2 & 31.6 & 73.6 & \textbf{29.9} & 53.1 & 64.2 & 10.1 & 39.4 \\ \cline{2-14}
 & \firstQ & 53.4 & 7.3 & \textbf{36.7} & 51.6 & 7.7 & 36.3 & 55.8 & 7.1 & 36.9 & 65.4 & \textbf{11.6} & \textbf{41.7} \\ \cline{2-14}
   & \firstK & 59.8 & 11.0 & 37.8 & 60.3 & 12.3 & 38.9 & 69.0 & 7.7 & 44.9 & 47.2 & 4.4 & 33.9 \\ \cline{2-14}
   & \firstV & 49.4 & 6.5 & 29.9 & 55.0 & 7.0 & 35.6 & 58.7 & 10.5 & 40.1 & 37.8 & 3.3 & 24.6 \\ \cline{2-14}
  & \firstAout & 36.6 & 3.3 & 25.5 & 35.7 & 4.3 & 25.3 & 60.6 & 17.5 & 44.6 & 50.0 & 3.8 & 30.1 \\ \cline{2-14}
  & \firstFin & 41.8 & 7.8 & 30.2 & 41.3 & 3.5 & 31.3 & \textbf{80.2} & 23.3 & 52.3 & 51.3 & 7.8 & 35.3 \\ \cline{2-14}
  & \firstFout & 46.2 & 5.1 & 29.6 & 44.8 & 7.1 & 30.4 & 79.0 & 24.8 & \textbf{55.9} & 59.4 & 5.4 & 35.7 \\ \bottomrule
\end{tabular}
}
\caption{The comparison of reconstruction attacks using different gradient modules on three datasets and four Transformer-based models ($B = 1$).}
\label{table:more_models}
\end{table*}

We observe that attacks on all four models achieved decent performance. TinyBERT, the smaller variant, exhibited the most vulnerabilities across the three datasets, while the fine-tuned version, BERT$_{\text{FT}}$, achieves performance comparable to the pre-trained version, BERT$_{\text{BASE}}$.

\newpage
\subsection{Attack Performance on Batches with Different Sizes}
\label{sec:sub_appendix_3}
We also test the attack performance on batches with different sizes, \ie $B = 1, 2, 4$. The results are presented in Table~\ref{table:more_batch_size}. We observe that employing gradients from intermediate modules achieved performance comparable to the baseline method, based on varying batch sizes.
\begin{table*}[!htbp]
\centering
\begin{tabular}{cc|ccc|ccc|ccc}
\toprule
\multirow{2}{*}{\textbf{Method}} & \multirow{2}{*}{\textbf{Gradient}} & \multicolumn{3}{c|}{\textbf{B=1}} & \multicolumn{3}{c|}{\textbf{B=2}} & \multicolumn{3}{c}{\textbf{B=4}} \\ \cline{3-11}
 &  &  R-1 & R-2 & R-L & R-1 & R-2 & R-L & R-1 & R-2 & R-L    \\ \bottomrule
\multicolumn{11}{c}{\textbf{Part A : \cola}}  \\ \toprule
LAMP & \alGrads & \textbf{89.9} & \textbf{58.0} & 80.0 & 71.2 & 23.1 & 58.9 & 50.3 & 11.5 & 43.8 \\ \midrule
\multirow{6}{*}{Ours} & \alTF & 87.8 & 54.2 & \textbf{80.4} & \textbf{76.3} & \textbf{41.2} & \textbf{66.3} & \textbf{54.3} & \textbf{13.8} & \textbf{46.5} \\ \cline{2-11}
  & \firstTF & 67.0 & 17.9 & 58.7 & 57.5 & 9.1 & 48.6 & 43.2 & 4.7 & 38.9 \\ \cline{2-11}
 & \firstQ & 79.9 & 26.6 & 65.0 & 45.7 & 7.5 & 40.7 & 39.3 & 4.5 & 35.9 \\ \cline{2-11}
  & \firstAout & 66.4 & 18.3 & 57.0 & 46.5 & 7.0 & 41.8 & 36.4 & 1.4 & 33.9 \\ \cline{2-11}
  & \firstFin & 59.2 & 15.9 & 54.0 & 49.1 & 5.7 & 42.3 & 36.5 & 4.4 & 35.0 \\ \cline{2-11}
  & \firstFout & 71.7 & 18.0 & 58.4 & 61.6 & 20.1 & 55.3 & 40.3 & 5.8 & 35.2 \\ \bottomrule
 \multicolumn{11}{c}{\textbf{Part B : \sst}}  \\ \toprule
LAMP & \alGrads & 92.6 & 74.5 & 85.6 & 73.0 & 44.7 & 69.8 & \textbf{63.9} & \textbf{29.2} & \textbf{59.5} \\ \midrule
\multirow{6}{*}{Ours} & \alTF & \textbf{94.9} & \textbf{77.4} & \textbf{87.1} & \textbf{78.1} & \textbf{48.5} & \textbf{71.8} & 63.5 & 25.5 & 57.7 \\ \cline{2-11}
  & \firstTF & 85.8 & 71.7 & 82.1 & 55.1 & 19.2 & 54.5 & 43.3 & 4.0 & 42.2 \\ \cline{2-11}
 & \firstQ & 84.3 & 58.0 & 79.7 & 58.7 & 20.3 & 55.8 & 40.6 & 5.1 & 40.1 \\\cline{2-11}
  & \firstAout & 76.0 & 52.5 & 73.9 & 53.9 & 8.0 & 50.9 & 48.9 & 10.8 & 47.5 \\ \cline{2-11}
  & \firstFin & 68.1 & 42.5 & 66.9 & 51.4 & 16.4 & 50.6 & 39.5 & 7.3 & 39.0 \\\cline{2-11}
  & \firstFout & 83.9 & 64.8 & 80.6 & 60.5 & 14.9 & 56.3 & 47.9 & 7.3 & 45.9 \\\bottomrule
 \multicolumn{11}{c}{\textbf{Part C : \rotten}}  \\ \toprule
LAMP & \alGrads & 59.2 & \textbf{10.3} & 34.6 & 31.5 & 4.1 & 22.7 & 21.3 & 1.1 & 19.0 \\ \midrule
\multirow{6}{*}{Ours} & \alTF & \textbf{61.6} & 6.4 & 35.9 & \textbf{36.6} & \textbf{4.5} & \textbf{25.4} & 22.9 & 0.7 & 19.6 \\ \cline{2-11}
  & \firstTF & 52.5 & 6.2 & 34.8 & 24.9 & 0.7 & 20.2 & 23.1 & \textbf{1.4} & 19.0 \\ \cline{2-11}
 & \firstQ & 53.4 & 7.3 & \textbf{36.7} & 29.8 & 2.8 & 23.0 & \textbf{24.7} & 0.6 & \textbf{20.2} \\ \cline{2-11}
  & \firstAout & 36.6 & 3.3 & 25.5 & 25.6 & 1.0 & 21.3 & 20.3 & 0.1 & 17.7 \\ \cline{2-11}
  & \firstFin & 41.8 & 7.8 & 30.2 & 21.1 & 2.0 & 17.3 & 20.0 & 0.2 & 18.0 \\ \cline{2-11}
  & \firstFout & 46.2 & 5.1 & 29.6 & 25.5 & 1.0 & 19.4 & 24.1 & 0.4 & 20.0 \\ \bottomrule
\end{tabular}
\caption{The comparison of attack performance across different batch sizes (\ie $B=1, 2, 4$) using BERT$_{\text{BASE}}$ model on three datasets.}
\label{table:more_batch_size}
\end{table*}

\newpage
\section{Examples of Reconstruction}
\label{sec:appendix_b}

Some reconstructed examples are presented in Table~\ref{table:example}. For each example, we provide results using parameter gradients from different modules: the baseline LAMP method, which utilizes gradients from all layers, and our method, which employs gradients specifically from the 9th Transformer layer, the 5th Attention Key module, and the 7th FFN Output module, respectively. We observe that the reconstruction performance of using only one Transformer layer or even a single module could recover inputs with coherent pieces. Their performance was comparable to the baseline method using full gradients.

\begin{table}[ht]
\centering
\renewcommand{\arraystretch}{1.5}
\resizebox{\columnwidth}{!}{%
\begin{tabular}{cccl}
\toprule
\textbf{Dataset} &  & \textbf{Gradients} & \multicolumn{1}{c}{\textbf{Sequence}} \\
\bottomrule
\multirow{5}{*}{CoLA} & \textcolor{lightblue}{Reference} &   & \textcolor{lightblue}{harriet alternated folk songs and pop songs together.} \\
\cline{2-4}
 & LAMP & $\Delta \mW_{all}$  & \colorbox{lightgreen}{harriet alternated folk songs and} \colorbox{lightyellow}{alternate} \colorbox{lightgreen}{songs together.} \\
\cline{2-4}
 & \multirow{3}{*}{Ours} & $\Delta \mW_{t,9}$  & \colorbox{lightgreen}{harriet alternated folk songs and pop songs together.} \\
\cline{3-4}
 & & $\Delta \mW_{k,5}$  & \colorbox{lightgreen}{harriet alternated} \colorbox{lightyellow}{songs} \colorbox{lightgreen}{and pop} \colorbox{lightgreen}{folk songs} \colorbox{lightgreen}{together.} \\
\cline{3-4}
 & & $\Delta \mW_{p,7}$  & \colorbox{lightyellow}{pop} \colorbox{lightyellow}{harriet} \colorbox{lightgreen}{songs and} \colorbox{lightgreen}{folk songs} \colorbox{lightyellow}{alternate} \colorbox{lightyellow}{together} him. \\
\bottomrule
\multirow{5}{*}{SST-2} & \textcolor{lightblue}{Reference} &   & \textcolor{lightblue}{hide new secretions from the parental units} \\
\cline{2-4}
 & LAMP & $\Delta \mW_{all}$  & \colorbox{lightyellow}{hide} \colorbox{lightyellow}{secretions} \colorbox{lightgreen}{from the} \colorbox{lightyellow}{new} \colorbox{lightgreen}{parental units} \\
\cline{2-4}
 & \multirow{3}{*}{Ours} & $\Delta \mW_{t,9}$  & hideions \colorbox{lightyellow}{hide} \colorbox{lightgreen}{from the} \colorbox{lightyellow}{new} \colorbox{lightgreen}{parental units} \\
\cline{3-4}
 & & $\Delta \mW_{k,5}$  & \colorbox{lightyellow}{units} \colorbox{lightyellow}{hide} \colorbox{lightyellow}{the} \colorbox{lightyellow}{secretions} \colorbox{lightyellow}{parental} \colorbox{lightgreen}{parental units} \\
\cline{3-4}
 & & $\Delta \mW_{p,7}$  & \colorbox{lightyellow}{units} \colorbox{lightyellow}{hide} secret \colorbox{lightgreen}{from the} \colorbox{lightyellow}{new} \colorbox{lightgreen}{parental units} \\
\bottomrule
 & \textcolor{lightblue}{Reference} &   & \textcolor{lightblue}{it will delight newcomers to the story and those who know it from bygone days .} \\
\cline{2-4}
\multirow{2}{*}{Rotten} & LAMP & $\Delta \mW_{all}$  & \colorbox{lightyellow}{it} \colorbox{lightyellow}{story} \colorbox{lightgreen}{will delight newcomers} \colorbox{lightgreen}{and those} by \colorbox{lightyellow}{those} \colorbox{lightyellow}{from} teeth \_ \colorbox{lightyellow}{story} \colorbox{lightyellow}{days} still \colorbox{lightgreen}{know it}. \\
\cline{2-4}
 & \multirow{3}{*}{Ours} & $\Delta \mW_{t,9}$  & \colorbox{lightgreen}{it will delight newcomers to} its \colorbox{lightyellow}{story} \colorbox{lightyellow}{who} already \colorbox{lightyellow}{know} \colorbox{lightyellow}{from} the word \colorbox{lightyellow}{and} prefer \colorbox{lightyellow}{those} \colorbox{lightgreen}{days.} \\
\cline{3-4}
Tomatoes  & & $\Delta \mW_{k,5}$  & \colorbox{lightyellow}{it} favored \colorbox{lightgreen}{newcomers to} \colorbox{lightyellow}{know} \colorbox{lightgreen}{the story and} \colorbox{lightyellow}{will} by \colorbox{lightgreen}{those who} \colorbox{lightyellow}{delight} \colorbox{lightyellow}{from} \colorbox{lightyellow}{it} daysgon. \\
\cline{3-4}
 & & $\Delta \mW_{p,7}$  & \colorbox{lightyellow}{it} \colorbox{lightyellow}{story} \colorbox{lightyellow}{to} \colorbox{lightgreen}{delight newcomers} \colorbox{lightyellow}{and} \colorbox{lightyellow}{delight} \colorbox{lightyellow}{those} \colorbox{lightgreen}{it will} \colorbox{lightyellow}{know} \colorbox{lightyellow}{from} \colorbox{lightyellow}{the} daysgone grandchildren. \\
\bottomrule
\end{tabular}%
}
\caption{Text reconstruction results for several examples from three datasets by using different gradient modules (for the BERT$_{\text{BASE}}$ model with a batch size of 1). Correct single words are highlighted in \colorbox{lightyellow}{yellow}, while correct phrases (consisting of more than one word) are highlighted in \colorbox{lightgreen}{green}.}
\label{table:example}
\end{table}

\end{document}